\documentclass[letterpaper,10pt,conference]{ieeeconf}
\usepackage{amsmath}
\usepackage{graphicx}
\usepackage{placeins}

\usepackage{dblfloatfix}
\usepackage{balance}
\IEEEoverridecommandlockouts
\title{\LARGE \bf
Hybrid Residual Reinforcement Learning for Contact-Rich Robotic Book Insertion}

\author{%
  Tianyuan Liu,
  Rutherford Agbeshi Patamia,
  Benjamin Champion,
  Akansel Cosgun,
  Richard Dazeley$^*$%
  \thanks{$^*$All authors are with Deakin University, Australia.
  Email: \{tianyuan.liu, r.patamia, benjamin.champion, akan.cosgun, richard.dazeley\}@deakin.edu.au}
}

\begin{document}
\maketitle
\thispagestyle{empty}
\pagestyle{empty}

\begin{abstract}
Placing a grasped book into a tight shelf is a compact but difficult contact-rich control problem: millimetre-scale pose error can turn a geometrically valid approach into jamming, failed release, or incomplete seating. We study this final phase after grasp acquisition and global approach, and ask how control authority should be divided between known geometry and learned behaviour. Our method retains a nominal task-space controller for structured insertion and seating, while residual PPO supplies bounded local corrections and decides when to release. Only the brief open--retreat--reclose transition is scripted. For the final policy used on hardware, a deployment-matched simulation evaluation over 512 fixed conditions yields $98.50\%$ mean success (0.23 percentage-point sample SD) across three independent training runs, compared with $37.89\%$ for nominal control. On the physical xArm7, 60 trials over 30 matched conditions show the same qualitative advantage: residual control raises success from $26.7\%$ to $63.3\%$, reduces failures from 22 to 11, and wins 13 of the 15 matched conditions in which the two controllers differ. Robustness tests show that performance remains above $87\%$ under initialization perturbations up to $1.5\times$, while very tight clearances expose the geometric limit of local correction. These results support a hybrid design in which geometry preserves reliable task structure and learning is concentrated on the contact-sensitive behaviour that fixed rules handle poorly.
\end{abstract}

\section{Introduction}
Bookshelf insertion is deceptively simple. Once a book is grasped and brought in front of a shelf, the remaining motion is only a few centimetres, yet this final phase can be harder than the global approach. A book that is slightly displaced or rotated may contact a neighbour at the shelf mouth, rotate under friction, fail to reach a release-ready configuration, or remain insufficiently seated after the gripper opens. These failures arise even when the nominal target pose is kinematically reachable. The problem is therefore not only where the book should end, but how the robot should move through uncertain contact on the way there.

\begin{figure}[t]
    \centering
    \includegraphics[width=0.96\columnwidth]{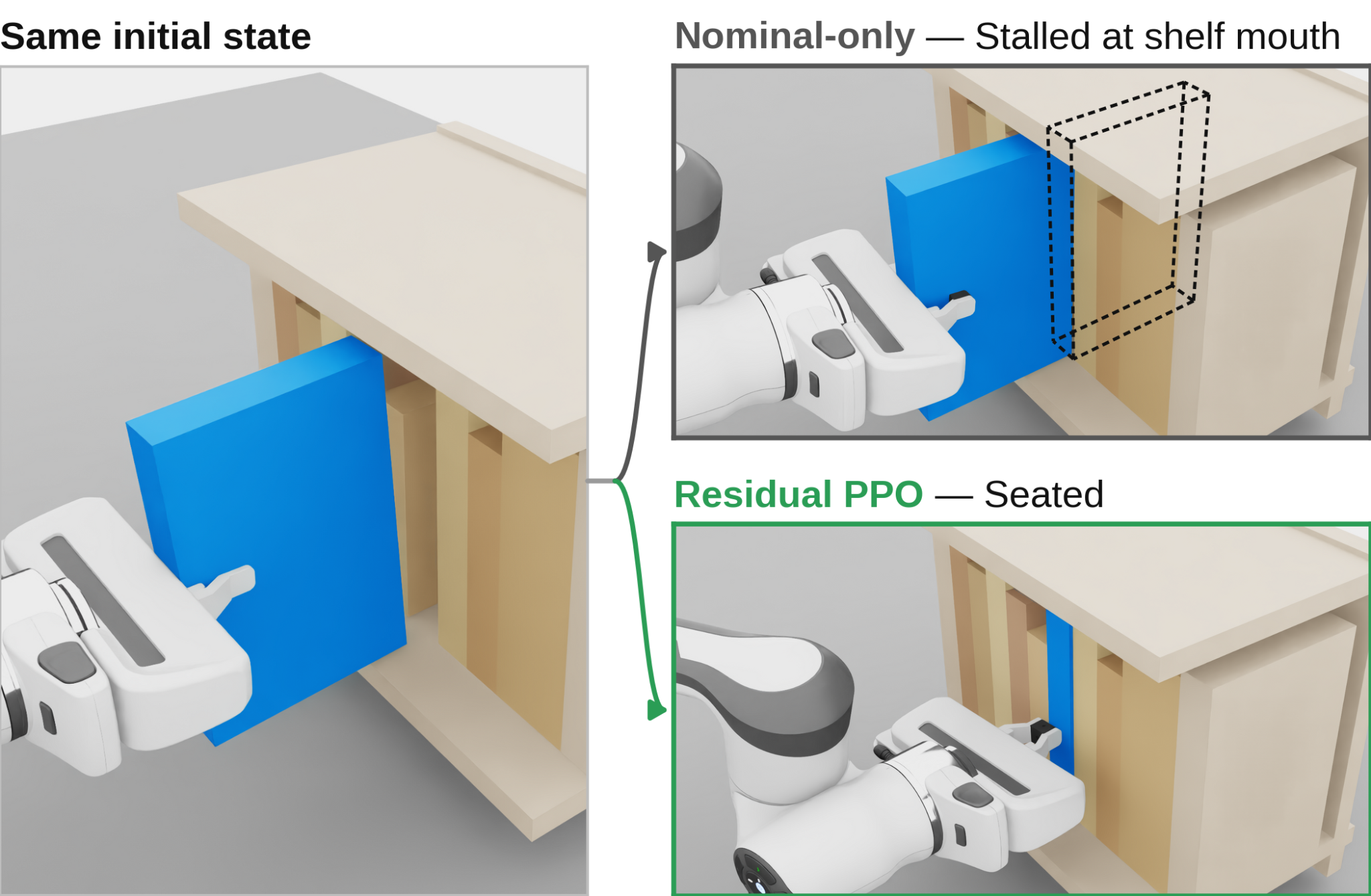}
    \caption{Same initial state, two outcomes in simulation. Nominal geometric control reaches the shelf mouth but can stall without releasing, while adding residual PPO to the same controller enables insertion, release, and final seating. The dashed outline marks the fully seated target pose.}
    \label{fig:money_shot}
\end{figure}

Purely geometric control is attractive because the task contains strong known structure: the shelf defines a clear insertion axis, lateral and angular alignment can be measured, and post-release seating has a predictable direction. However, a fixed controller must encode contact behaviour implicitly through hand-tuned gains and thresholds, and becomes brittle when grasp error, local friction, or neighbouring-book geometry differs from its nominal assumptions. At the other extreme, learning the full task-space control policy without a geometric prior requires the policy to rediscover task progression that is already available analytically, which can make training unnecessarily difficult.

This motivates a narrower question: \emph{which parts of tight book insertion should remain geometric, and which should be learned?} We retain a nominal task-space controller for structured progress and learn only a bounded residual correction together with the release decision. The policy is active both while the book is grasped and during the post-release push. Only a short mechanical open--retreat--reclose transition is scripted. The task begins from a feasible pre-insertion handoff, so grasp acquisition, global planning, and raw visual perception are deliberately outside the scope of this work.

The paper makes three contributions:
\begin{itemize}
    \item We formulate final-phase bookshelf insertion as a hybrid sequence spanning grasped insertion, learned release, and post-release seating, rather than treating successful arrival at an intermediate pose as task completion.
    \item We introduce a phase-aware control-authority decomposition in which geometry supplies interpretable task progress, while residual PPO provides bounded contact-sensitive corrections and determines the release decision across grasped insertion and post-release seating.
    \item We evaluate the decomposition with matched multi-run simulation, robustness and training analyses, and a paired 60-trial xArm7 study. On hardware, residual control improves success by $36.7$ percentage points and recovers 13 of the 22 conditions in which nominal control fails.
\end{itemize}

\section{Related Work}
\subsection{Book Insertion and Constrained Placement}
Robotic manipulation of books has been studied through task-specific insertion strategies, shelf rearrangement, and contact-aware planning. Early work examined book arrangement and insertion mechanics~\cite{nakajima2011study1book1arrangement}. More recent systems combine perception and multi-stage manipulation for rearranging bookshelves~\cite{sygo2023multistage}, while Yang \emph{et al.} use model-based contact planning for book insertion~\cite{yang2025planning}. Reinforcement learning has also appeared in a Visual Bookshelf task~\cite{singh2019endtoend}. These works establish bookshelves as a meaningful constrained-manipulation domain, but do not address our specific design question: how to divide control authority between a geometric insertion controller and a learned local correction policy during the final contact-rich phase.

The problem is related to constrained object placement more broadly. Prior work considers task-driven placement, reactive placement, and learned pose or arrangement generation~\cite{mitash2020task1driven,lach2023placing,kreis2023reactive,kreis2024compact,zhao2025anyplace}. Our setting is deliberately downstream of candidate generation and grasp selection. The target book is already grasped, the shelf slot is known, and a global planner has delivered the robot to pre-insertion. This isolates whether learned local control can improve execution robustness once geometric uncertainty turns into physical contact.

\subsection{Contact-Rich Manipulation and Residual Learning}
Contact-rich manipulation is sensitive to modelling error, compliance, and unobserved contact state~\cite{suomalainen2022survey,elguea2023review}. Learning-based approaches have therefore been explored for insertion, assembly, and adaptive contact skills~\cite{schoettler2020meta,padalkar2024guiding}. Residual reinforcement learning provides a natural compromise when a useful conventional controller already exists: a policy learns a correction rather than the entire control law~\cite{johannink2019residual,ranjbar2021residual1contactrich}. Related work has combined residual learning with demonstrations or high-precision manipulation priors~\cite{alakuijala2021residualfromdemonstration,davchev2022residual,ankile2407imitation,zhang2024residualrlmethod}.

Building on this line of work, we study bookshelf insertion as a phase-structured hybrid control problem. The residual policy not only modifies the nominal insertion command, but also determines when to release the book and resumes control during post-release seating. We evaluate this decomposition against nominal-only and PPO-only control in simulation, and further test whether the same task-space interface transfers to a different 7-DoF arm.

\section{Method}
\label{sec:method}
\subsection{Problem Formulation and Phase Structure}
Each episode begins with the target book already grasped and the robot at a feasible pre-insertion pose. We define a local coordinate frame at the target slot, with \(x\) pointing from the shelf mouth toward the back of the shelf, \(y\) spanning the slot laterally, and \(z\) pointing upward. Book pose errors and controller commands are expressed in this frame.

\begin{figure}[t]
    \centering
    \includegraphics[width=\columnwidth]{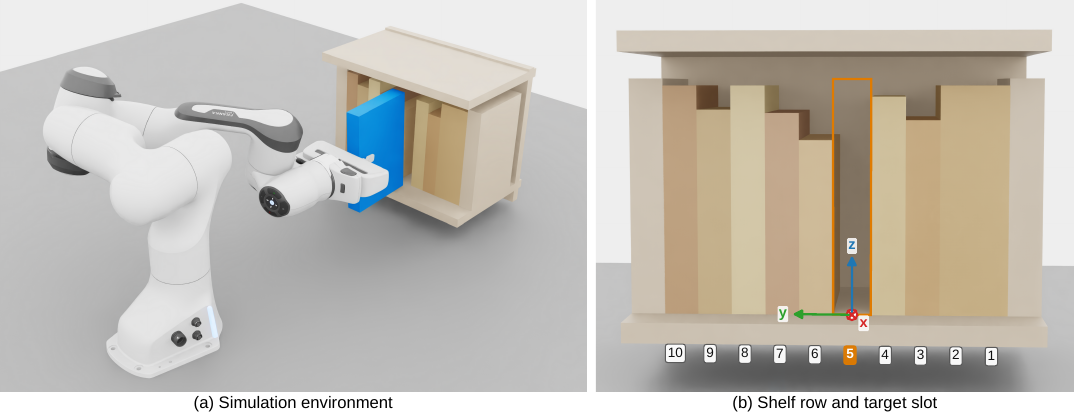}
    \caption{Simulation environment and bookshelf arrangement.
    (a) Franka Panda performing the final bookshelf-insertion task
    with the target book initialized at a feasible pre-insertion pose.
    (b) Frontal view of the simulated shelf showing the ten possible
    insertion locations and an example target slot. The local slot
    frame defines \(x\) into the shelf, \(y\) along the shelf row,
    and \(z\) upward.}
    \label{fig:sim_setup}
\end{figure}

The insertion task consists of three control phases: grasped insertion, a deterministic release transition, and post-release seating. The policy controls the grasped insertion and post-release seating phases. During the brief release transition, the gripper executes a fixed open--retreat--reclose sequence before policy control resumes.

\subsubsection{Observation Space}

The policy operates on geometric task state rather than images or force measurements. It receives the 12-dimensional observation summarized in Table~\ref{tab:observation}. The observation captures insertion progress, alignment with the target slot, the relative pose between the tool and book, gripper state, and the current task phase.

\begin{table}[t]
\centering
\caption{Residual policy observation space.}
\label{tab:observation}
\footnotesize
\begin{tabular}{c l l}
\hline
\textbf{Dim.} & \textbf{Observation} & \textbf{Scale / Encoding} \\
\hline
0 & Task phase & $0$, $0.5$, $1$ \\
1--2 & Insertion depth and remaining depth & $0.08$ m each \\
3--5 & Lateral, vertical, and yaw errors &
$0.05$ m, $0.05$ m, $30^\circ$ \\
6--8 & Tool-to-book displacement $(x,y,z)$ & $0.25$ m each \\
9 & Gripper opening & Normalized \\
10--11 & Book spine direction $(x,y)$ & Unit-vector components \\
\hline
\end{tabular}
\end{table}

The task-phase value is \(0\) during grasped insertion, \(0.5\) during the deterministic release transition, and \(1\) during post-release seating. Geometric quantities are divided by the scales listed in Table~\ref{tab:observation} and clipped to \([-1,1]\). VecNormalize is then applied, with normalized observations clipped to \([-10,10]\).

\subsubsection{Action Space}

The policy outputs a six-dimensional action
\begin{equation}
a_t =
[a_x,a_y,a_z,a_{\mathrm{yaw}},a_{\mathrm{pitch}},a_g].
\end{equation}
The first five components specify incremental task-space motion corrections. The translational components adjust forward insertion, lateral alignment, and height, while the rotational components adjust yaw and pitch. These actions are all \textbf{local corrections} rather than absolute target poses.

A unit action corresponds to \(2.0\,\mathrm{mm}\), \(1.0\,\mathrm{mm}\), and \(1.5\,\mathrm{mm}\) along the local \(x\), \(y\), and \(z\) axes, respectively, and \(0.35^\circ\) and \(0.30^\circ\) in yaw and pitch. The sixth component \(a_g\) controls release. A release is requested when \(a_g>0.5\).

\subsubsection{Episode Completion}

Successful insertion requires more than moving the front of the book through the shelf opening. The book must be released and subsequently seated during the post-release phase.

In simulation, success requires the book to be released and stably seated inside the target slot. The book must be inserted sufficiently deeply toward the shelf back, remain fully within the lateral slot boundaries, and maintain acceptable vertical alignment, yaw, and upright orientation. These conditions must hold for four consecutive control updates before the episode timeout. Simulation success requires four consecutive updates with the rear face no more than \(12\,\mathrm{mm}\) outside the shelf mouth, the leading face within \(55.2\,\mathrm{mm}\) of the slot back, full lateral containment with \(1.5\,\mathrm{mm}\) numerical tolerance, vertical error below \(15\,\mathrm{mm}\), yaw error below \(8^\circ\), and upright-axis dot product at least \(0.85\). Episodes time out after \(10\,\mathrm{s}\).

\subsection{Nominal and Residual Control Authority}
Let $\Delta a_t^{\mathrm{nom}}$ denote the nominal task-space increment and $\Delta a_t^{\mathrm{res}}$ the learned residual. During insertion and post-release push,
\begin{equation}
    \Delta a_t = \mathrm{clip}\!\left(\Delta a_t^{\mathrm{nom}}+\Delta a_t^{\mathrm{res}}\right),
    \label{eq:residual_fusion}
\end{equation}

\begin{figure}[h!]
    \centering
    \includegraphics[
    width=0.5\columnwidth,
    trim=20 20 20 10,
    clip
]{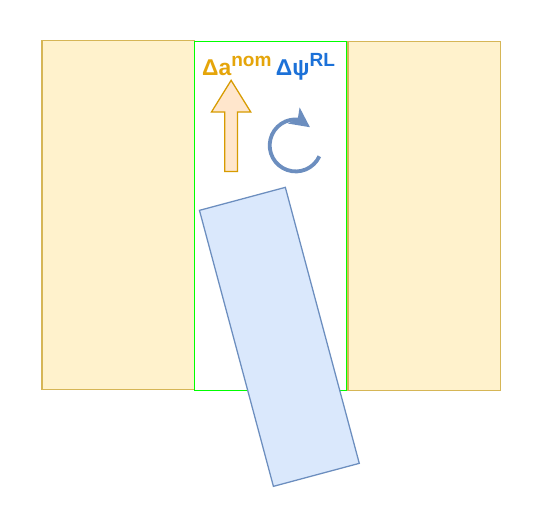}
    \caption{Conceptual residual correction. The nominal controller provides structured task-space progress, while the learned policy supplies a bounded local correction. A rotational correction is shown here for illustration.}
    \label{fig:residual_correction}
\end{figure}

where clipping enforces the same per-step task-space limits used during training and deployment. The decomposition therefore retains the geometric controller as a structured local prior while giving the policy limited authority to modify its command.

During grasped insertion, the nominal controller continuously corrects lateral position, height, yaw, and pitch relative to the target slot. Forward motion is allowed only when the book is sufficiently well aligned, preventing the controller from forcing the book deeper into the shelf while large pose errors remain. The commanded forward step is also reduced as the book moves deeper into the slot. After release, the nominal controller continues to provide forward pushing and geometric alignment to seat the book. This controller therefore provides a structured and conservative geometric prior, but it cannot adapt its corrections to contact-induced deviations that are not captured by the geometric error alone.

Residual PPO acts at every learned control step and can increase, decrease, or redirect the nominal increment within its bounded action range. The policy also determines when to request release. After a release request, a short deterministic sequence opens the gripper, retreats, recloses it, and prepares the empty gripper for the seating phase. Learning then resumes during post-release push, so final seating is not purely scripted. Real deployment evaluates the same task-space policy at $20\,\mathrm{Hz}$ and realizes the resulting targets through the xArm Servo/velocity layer.

\begin{figure*}[t]
    \centering
    \begin{minipage}[c]{0.69\textwidth}
        \centering
        \includegraphics[width=\linewidth]{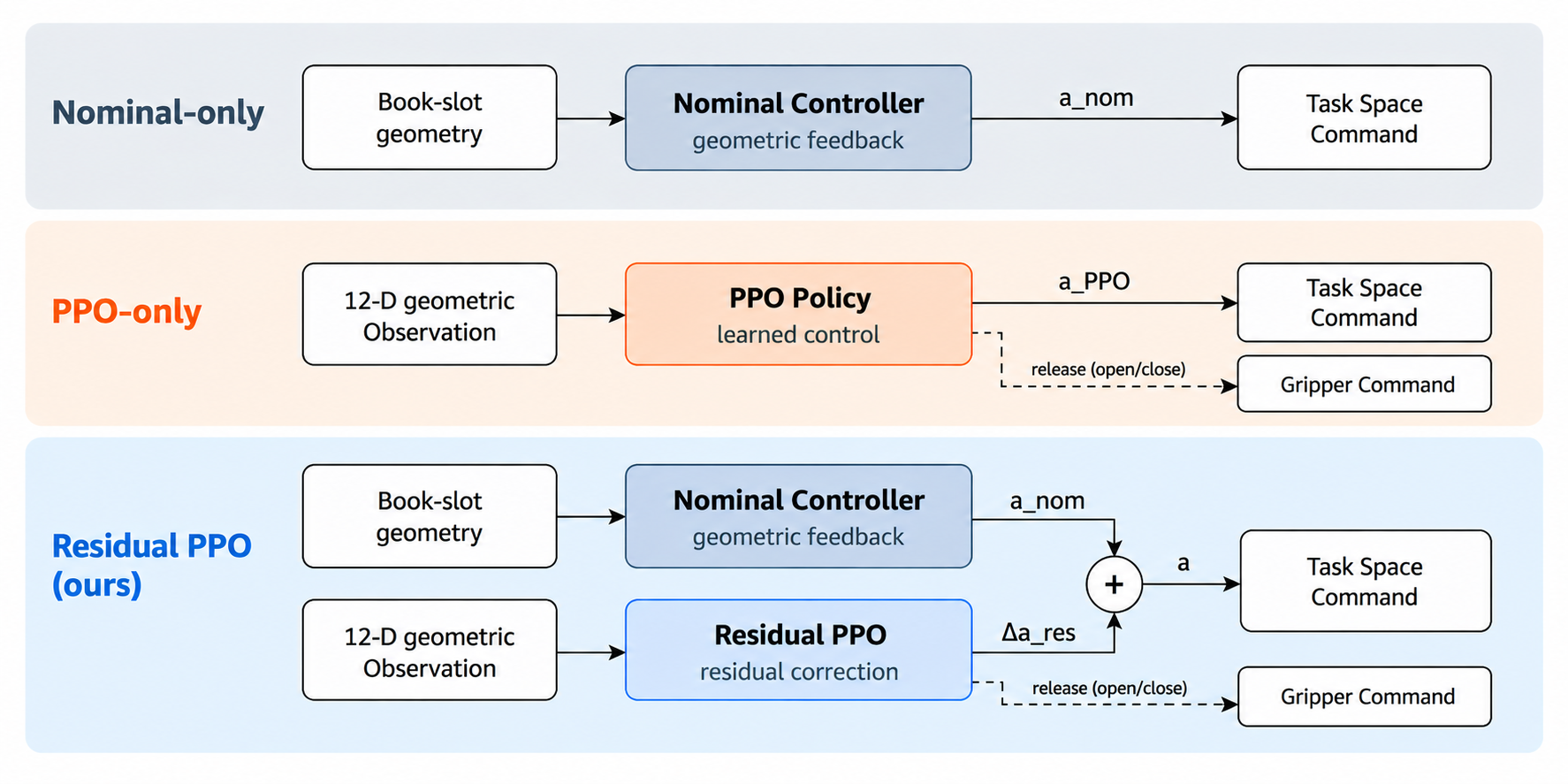}\\[-0.5mm]
        {\small (a) Controller formulations}
    \end{minipage}\hfill
    \begin{minipage}[c]{0.285\textwidth}
        \centering
        \includegraphics[width=\linewidth]{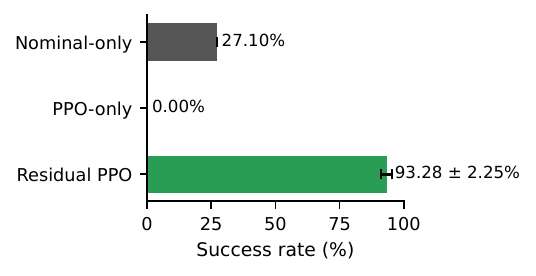}\\[-0.5mm]
        {\small (b) Controller comparison at 3 mm}
    \end{minipage}
    \caption{Controller formulations and simulation comparison. (a) Nominal-only follows geometric feedback, PPO-only directly controls the task-space action, and Residual PPO adds a bounded learned correction to nominal control while also learning the release request. (b) At $3\,\mathrm{mm}$ slot clearance, all three formulations are evaluated with the final evaluator on the same frozen 2,000-condition scenario bank. Nominal-only achieves $27.10\%$ success, PPO-only achieves $0\%$ across three independently trained policies, and Residual PPO achieves $93.28\pm2.25\%$ mean success across three independent training runs.}
    \label{fig:main_comparison}
\end{figure*}

\subsection{Training Objective and Curriculum}
The final residual policy is trained with PPO in Isaac Lab using the settings summarized in Table~\ref{tab:training_settings}.

\begin{table}[t]
\centering
\caption{Training and deployment settings of the final residual policy.}
\label{tab:training_settings}
\scriptsize
\setlength{\tabcolsep}{3pt}
\renewcommand{\arraystretch}{1.08}
\begin{tabular}{p{0.43\columnwidth}p{0.49\columnwidth}}
\hline
\textbf{Setting} & \textbf{Value} \\
\hline
Algorithm / implementation & PPO / Stable-Baselines3 2.8.0 \\
Parallel environments & 256 \\
Physics / control rate & 120 / 60 Hz \\
Observation / action & 12-D / 6-D \\
Actor / critic & $12$-$256$-$256$-$6$ / $12$-$256$-$256$-$1$ \\
Activation & ReLU \\
Rollout / batch / epochs & 32 / 8192 / 10 \\
Learning rate & $10^{-4}$ \\
$\gamma$ / GAE $\lambda$ / PPO clip & 0.99 / 0.95 / 0.20 \\
Entropy / value coefficient & 0.003 / 1.0 \\
Maximum gradient norm & 1.0 \\
Observation normalization / clip & yes / 10 \\
Reward normalization & no \\
Training clearance & $3\,\mathrm{mm}$ \\
Residual-action $L_2$ coefficient & 0.01 \\
\hline
\end{tabular}
\end{table}
The reward encourages insertion progress while penalizing lateral, vertical, and yaw misalignment. The same progress objective is retained during post-release seating, with a smaller lateral penalty to permit contact-driven adjustment. Successful completion receives a terminal bonus, book drops are penalized, and a small \(L_2\) penalty discourages unnecessarily large residual actions.

Training uses a two-stage curriculum at a fixed \(3\,\mathrm{mm}\) total lateral clearance. The first stage uses relatively small grasp and reset perturbations, while the second stage broadens the initial position, orientation, and joint perturbations for greater robustness. The transition occurs approximately midway through training. Shelf layouts contain ten possible insertion locations with randomized target gaps and neighbouring-book configurations to expose the policy to varying local contact geometry.

\section{Experimental Setup}
\subsection{Simulation Protocols}
Simulation uses Isaac Lab / Isaac Sim with a Franka Panda, a \(156\times236\times34\,\mathrm{mm}\) target book, \(120\,\mathrm{Hz}\) physics, and \(60\,\mathrm{Hz}\) control. We evaluate the method under three complementary protocols.

First, a \(3\,\mathrm{mm}\) controller comparison evaluates Nominal-only, PPO-only, and Residual PPO on the same frozen 2,000-condition scenario bank using the final evaluator. PPO-only and Residual PPO are each evaluated across three independently trained policies. This comparison is used to study the benefit of combining geometric control with learned residual corrections.

Second, robustness sweeps evaluate performance across different slot clearances and initialization perturbations. Nominal-only and Residual PPO are evaluated with the final evaluator using identical frozen scenario banks at each sweep point. Residual PPO reports the mean across three independently trained policies.

Third, a deployment-matched evaluation tests three independently trained final policies on the same 512 fixed initial conditions at \(3.84\,\mathrm{mm}\) clearance. Nominal control is evaluated on the identical set. This protocol most closely matches the policy configuration subsequently used for hardware experiments.
\subsection{Real-World Platform and Paired Protocol}

Hardware experiments use a 7-DoF xArm7 under ROS~2. The book is grasped before the insertion controller begins. Its pose is reconstructed from the measured TCP pose and a calibrated grasp-specific TCP-to-book transform, while a fixed reference-book calibration defines the target slot frame. MoveIt is used only to reach a feasible pre-insertion pose, with the book positioned approximately \(30\,\mathrm{mm}\) before the shelf mouth.

For each matched initial condition, we evaluate either \textbf{Nominal-only} control or \textbf{Residual PPO}, where the learned policy provides bounded task-space corrections to the same nominal controller described in Section~\ref{sec:method}. The resulting task-space commands are executed on the xArm7 through its Servo/velocity control layer. Detailed contact with neighbouring books is not represented in the motion-planning scene and is instead handled during insertion by the local controller.

\begin{figure}[t]
    \centering
    \includegraphics[width=0.90\columnwidth]{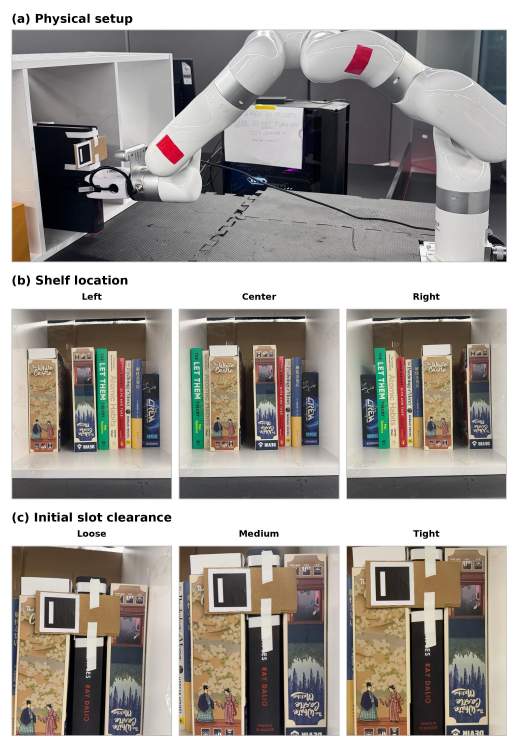}
    \caption{Real-world xArm7 setup and physical test factors. The shelf-location row shows the three tested parts of the shelf, while the physical-constraint row shows Loose, Medium, and Tight starting configurations. These photographs depict initial conditions rather than goal poses.}
    \label{fig:real_world_setup}
\end{figure}

The real-world evaluation contains 30 matched experimental conditions, summarized in Table~\ref{tab:realworld_protocol}. Each condition is executed once with Nominal-only control and once with Residual PPO, giving 60 trials in total. Controller order is randomized within each pair while the physical configuration and perturbation are held fixed. Physical shelf location is used as a blocking factor rather than as a fully crossed experimental factor.

\begin{table}[t]
\centering
\caption{Real-world evaluation design.}
\label{tab:realworld_protocol}
\footnotesize
\setlength{\tabcolsep}{3pt}
\begin{tabular}{p{0.27\columnwidth}p{0.63\columnwidth}}
\hline
\textbf{Factor} & \textbf{Setting} \\
\hline
Controllers &
Nominal-only, Residual PPO \\

Physical constraint &
Loose, Medium, Tight \\

Perturbations &
2 centred, $\pm3\,\mathrm{mm}$ lateral,
$\pm3^\circ$ yaw, and 4 combined lateral--yaw offsets \\

Shelf locations &
3 physical locations used as blocking assignments \\

Matched conditions &
30 \\

Trials &
60 total, one trial per controller for each matched condition \\
\hline
\end{tabular}
\end{table}

The ten perturbation conditions include two centred trials, positive and negative \(3\,\mathrm{mm}\) lateral offsets, positive and negative \(3^\circ\) yaw offsets, and the four sign combinations of lateral and yaw error. Across the experiment, all nine physical-constraint--location combinations are represented, with each constraint--perturbation condition assigned to one shelf location.

Loose, Medium, and Tight describe increasing effective physical insertion constraint rather than calibrated rigid-body clearances. Unlike simulation, the physical books can deform, compress, rotate, and shift under contact, so a single geometric gap measurement does not fully characterize insertion difficulty. We therefore use these labels as categorical physical configurations rather than assigning each regime a nominal clearance in millimetres.

During execution, the controller determines its release timing and the control logic manages the subsequent phase transitions and termination. These internal signals indicate when the controller considers the insertion sequence complete, but they are not used alone as the ground-truth success label. Final real-world success is determined from the physical outcome after the complete execution, including whether the book has been released and successfully seated. Controller states and perception-derived seating measurements are recorded as supporting diagnostics.

The final dataset contains one adjudicated outcome for every scheduled trial. Attempts invalidated by infrastructure or execution-system faults could be repeated, while a completed controller outcome closed the corresponding matched condition. Because each physical condition is evaluated with both controllers, the primary statistical comparison is paired. We use the exact two-sided McNemar test on discordant pairs and report marginal success rates for both controllers.

\section{Results}
\subsection{Does the Hybrid Decomposition Matter?}\label{sec:results_hybrid}
The deployment-matched evaluation provides the simulation result for the final policy. On 512 fixed test conditions at approximately \(3.8\,\mathrm{mm}\) clearance, the three independently trained residual policies achieve \(98.50\pm0.23\%\) mean success, compared with \(37.89\%\) for Nominal-only control.

Figure~\ref{fig:main_comparison} separately compares the three controller formulations at $3\,\mathrm{mm}$ clearance under the same final evaluator and frozen scenario bank. Nominal-only control achieves $27.10\%$ success, while PPO-only achieves $0\%$ across all three independently trained policies. Residual PPO reaches $93.28\pm2.25\%$. Under the same task formulation and evaluation conditions, retaining the geometric controller therefore substantially simplifies the learning problem.

\subsection{Robustness to Clearance and Initialization Error}
\begin{figure*}[t]
    \centering
    \includegraphics[width=0.80\textwidth]{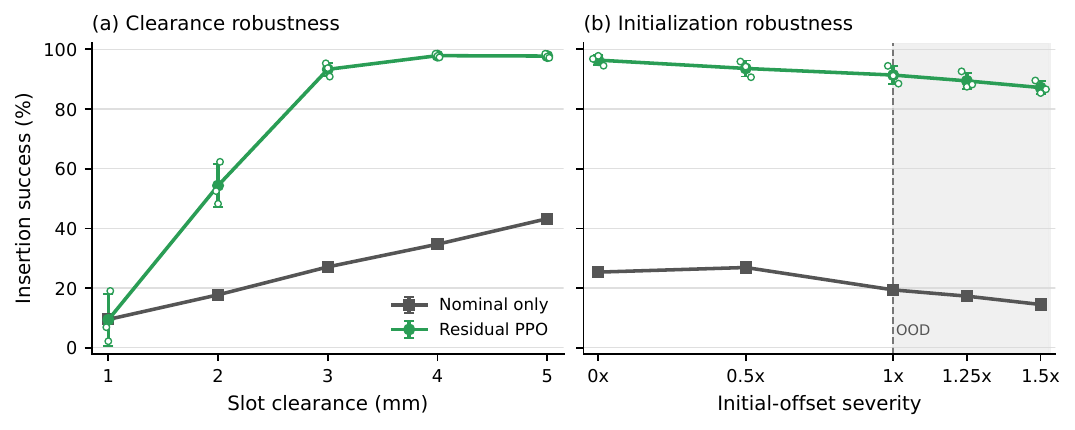}
    \caption{Robustness to slot clearance and initialization perturbation. Nominal-only and Residual PPO are evaluated with the final evaluator using identical frozen scenario banks at each sweep point. Residual PPO reports the mean and sample SD across three independently trained policies. Perturbation values beyond $1.0\times$ are outside the training range.}
    \label{fig:robustness}
\end{figure*}

Figure~\ref{fig:robustness} evaluates Nominal-only and Residual PPO under matched frozen conditions while varying slot clearance and initialization perturbations. Clearance is the dominant limitation. Residual PPO achieves \(93.28\%\) success at the \(3\,\mathrm{mm}\) training clearance and approximately \(98\%\) at \(4\)--\(5\,\mathrm{mm}\), but performance falls sharply as the slot becomes extremely tight, reaching \(54.35\%\) at \(2\,\mathrm{mm}\) and \(9.38\%\) at \(1\,\mathrm{mm}\). Variability across training runs also increases at the smallest clearances, indicating that near-contact geometric constraints become the principal failure mode.

In contrast, performance degrades gradually under increasing initialization perturbations. Success decreases from \(96.37\%\) without additional perturbation to \(91.37\%\) at the training-range level (\(1.0\times\)), and remains \(87.17\%\) at \(1.5\times\). Thus, the learned residual remains robust to initialization errors beyond those seen during training, while very small physical clearance presents a substantially harder constraint.

\subsection{What Happens During Learning?}
\begin{figure*}[t]
    \centering
    \includegraphics[width=0.78\textwidth]{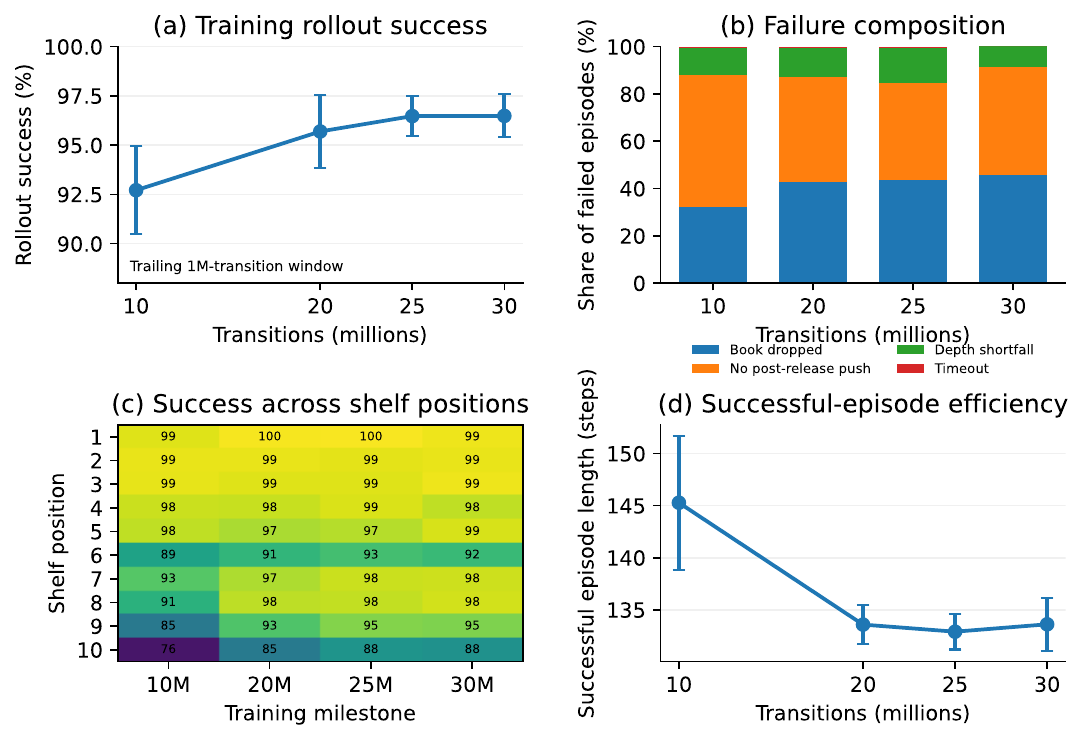}
    \caption{Training dynamics for the final residual-PPO configuration across three independent runs. (a) On-policy rollout success in the trailing one-million-transition window. (b) Failure composition conditioned on failed episodes. (c) Success across shelf positions. (d) Mean successful-episode length. Error bars show sample SD across runs where applicable.}
    \label{fig:training_dynamics}
\end{figure*}

Figure~\ref{fig:training_dynamics} summarizes on-policy rollout statistics from the three final-policy training runs. These curves describe training behaviour rather than held-out evaluation. Rollout success increases from \(92.71\%\) during the first third of training (10M transitions) to \(96.48\%\) at the final checkpoint (30M transitions), with little further improvement during the final portion of training. Successful episodes also become shorter, indicating that the policy completes the task more efficiently as training progresses.

Improvement is concentrated in the more difficult shelf positions. Locations that already perform well early in training remain stable, while the hardest position improves from roughly \(76\%\) success during the first third of training to about \(88\%\) by the final checkpoint. Failures associated with incomplete post-release seating and insufficient insertion depth also become less common. Although book drops account for a larger fraction of the remaining failures late in training, this occurs as the overall failure pool shrinks and does not indicate an increase in absolute drop frequency.

The curriculum begins with a narrower reset and grasp perturbation range before transitioning to a broader second-stage range around the middle of training. The policy then remains under this broader range for the remainder of training.

\begin{figure}[t]
    \centering
    \includegraphics[width=\columnwidth]{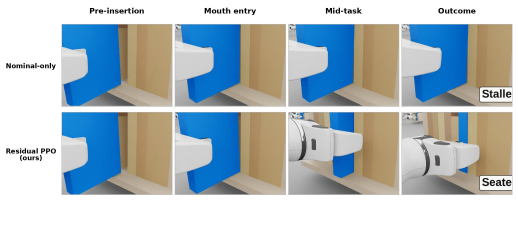}
    \caption{Representative matched execution at approximately $3.8\,\mathrm{mm}$ slot clearance. Both controllers start identically. Nominal-only stalls at the shelf mouth, whereas Residual PPO progresses through insertion and reaches a seated outcome.}
    \label{fig:matched_sequence}
\end{figure}

\subsection{Real-World Transfer}
\begin{table}[t]
\centering
\caption{Real-world success over 60 trials (30 matched conditions).}
\label{tab:real_results}
\small
\setlength{\tabcolsep}{3.5pt}
\begin{tabular}{lccc}
\hline
\textbf{Condition} & \textbf{Nominal} & \textbf{Residual PPO} & \textbf{Gain} \\
\hline
Overall & $8/30$ (26.7\%) & $19/30$ (63.3\%) & +36.7 pp \\
Loose   & $4/10$ (40\%) & $9/10$ (90\%) & +50 pp \\
Medium  & $3/10$ (30\%) & $7/10$ (70\%) & +40 pp \\
Tight   & $1/10$ (10\%) & $3/10$ (30\%) & +20 pp \\
\hline
\end{tabular}
\end{table}

Across the 30 matched conditions, Residual PPO increases end-to-end success from \(26.7\%\) (8/30) with Nominal-only control to \(63.3\%\) (19/30), reducing the number of failures from 22 to 11. The improvement is observed across all three physical-constraint categories. Success increases from \(40\%\) to \(90\%\) in Loose configurations, from \(30\%\) to \(70\%\) in Medium configurations, and from \(10\%\) to \(30\%\) in Tight configurations. Performance also improves at each of the three physical shelf locations, indicating that the overall gain is not driven by a single favourable region of the shelf.

The paired comparison provides the strongest evidence of the controller difference. Both methods succeed under six matched conditions and both fail under nine. Among the 15 conditions where their outcomes differ, Residual PPO succeeds in 13 and Nominal-only control in two. Thus, Residual PPO wins \(86.7\%\) of the direct disagreements and recovers 13 of the 22 conditions in which nominal control fails. An exact two-sided McNemar test confirms that the paired difference is significant (\(p=0.0074\)). Residual control therefore improves robustness substantially, although it does not dominate nominal control in every individual trial.

The Tight configuration remains challenging, with Residual PPO succeeding in only \(30\%\) of trials. These configurations provide very little free lateral motion and require careful alignment before insertion. During physical setup, even manual insertion required noticeably more precise alignment than in the Loose and Medium configurations. We therefore interpret Loose, Medium, and Tight as increasing categories of effective physical constraint rather than as calibrated numerical clearances. 

\subsection{Matched Qualitative Behavior}
Figure~\ref{fig:matched_sequence} provides a qualitative comparison from the same initial state. Under Nominal-only control, the book reaches the shelf mouth but makes little further insertion progress and never reaches the release stage. Residual PPO continues the insertion, requests release, and completes post-release seating. We therefore describe the nominal outcome as a stall, where forward progress effectively stops before release.

The sequence is intended to illustrate the difference in task progression rather than a specific corrective mechanism. The clearest visible distinction is continued insertion and successful completion under Residual PPO. Smaller lateral or pitch corrections are not reliably resolved in the printed frames, so the figure should not be interpreted as evidence of a unique correction direction or contact-force response.

\section{Discussion}
\subsection{Why the Hybrid Controller Helps}

The results highlight the benefit of combining geometric task structure with learned contact-sensitive adaptation. Bookshelf insertion provides useful geometric information, including a clear insertion direction, measurable alignment errors, and a natural direction for post-release seating. The nominal controller exploits this structure to provide consistent task progression, while the residual policy focuses on local corrections and release timing that become important during contact. This division is reflected in the controller comparison. Nominal-only control often progresses toward the slot but becomes brittle under tight contact and small modelling errors, whereas PPO-only fails to learn successful task completion under the same compact observation and training setup. Residual PPO preserves the structured geometric behaviour while adapting the motion locally when the nominal command is insufficient. Its improvement on the physical robot further suggests that this decomposition remains useful under real contact effects and simulation-to-hardware mismatch.

More broadly, the experiments illustrate a practical trade-off between hand-designed control and learning the full control policy from scratch. Explicitly engineering a geometric controller for every possible contact configuration can require substantial task-specific design and may still remain brittle to modelling errors. At the other extreme, asking a learned policy to discover both task progression and contact adaptation makes exploration considerably harder. The hybrid formulation avoids requiring either component to solve the entire problem. A relatively simple geometric controller supplies reliable task structure, while learning is concentrated on the contact-sensitive corrections and release decisions that are difficult to specify analytically. In this way, the method reduces the burden placed on both detailed controller engineering and learning from scratch.

\subsection{Simulation-to-Real Gap}
The deployment-matched simulation evaluation reaches \(98.5\%\), while the paired xArm7 experiment reaches \(63.3\%\). This gap reflects several physical effects that are simplified in simulation. In particular, neighbouring books are treated as fixed geometry in the simulator, so contact with them does not change the slot itself. On the real shelf, however, even a small contact can shift or rotate a neighbouring book, reducing the available opening and changing the contact geometry during the same insertion attempt. The simulated policy can therefore continue pushing against a geometrically unchanged slot in situations where the corresponding real-world contact may make the remaining insertion substantially harder or even infeasible.

Additional differences arise from grasp and shelf-frame calibration error, book compliance, friction, and small geometric imperfections. The real planning scene also does not model the detailed neighbouring-book contact geometry during insertion. The hardware result should therefore be interpreted as successful transfer of the hybrid control strategy rather than as evidence that simulation accurately predicts physical success probability.

The physical results show the same qualitative trend. Residual PPO improves success in the Loose, Medium, and Tight configurations, but performance remains lowest in the Tight condition, where little free lateral motion is available and neighbouring-book contact can easily disturb the opening. Because the physical books can move and deform, these categories should not be interpreted as calibrated numerical clearances. The controlled simulation sweep isolates geometric clearance more directly, showing a sharp performance drop below \(3\,\mathrm{mm}\). Together, these results suggest that residual correction can compensate for moderate pose and contact uncertainty, but it cannot recover when the available physical insertion space itself becomes too constrained.

\subsection{Limitations and Future Work}

This work focuses on the final contact-rich stage of bookshelf insertion. The book is assumed to have already been grasped and brought to a feasible pre-insertion pose, while perception, grasp selection, and global motion planning are handled outside the proposed controller. The policy also relies on geometric state estimates rather than visual, tactile, or force feedback. On hardware, book pose is reconstructed from the measured TCP pose using a calibrated rigid TCP-to-book transform, so any in-grasp slip is not directly observed. As a result, the controller can react to the geometric consequences of contact but does not directly observe the underlying contact forces or changes in the grasp transform.

The real-world evaluation is limited to one book geometry, one physical shelf setup, and three tested shelf locations. Simulation covers ten possible slot locations with randomized neighbouring-book layouts, but broader variation in book geometry, materials, shelf structure, and grasp configuration remains to be studied. In addition, neighbouring books are fixed in simulation, whereas real books can deform, shift, and rotate during contact. This difference becomes particularly important in tightly constrained insertions, where a small contact can change the available opening itself. The physical experiments therefore demonstrate transfer of the proposed control decomposition, but broader generalization across book sizes, materials, shelf geometries, and grasp configurations remains to be established.

Future work will extend the approach toward a more complete bookshelf-manipulation system by incorporating visual and contact feedback, broader object and shelf variation, and adaptation to changes in the local geometry during insertion. An important direction is to retain the geometric structure that simplifies learning while allowing the learned component to respond to richer physical interaction.

\section{Conclusion}

We studied the final contact-rich stage of robotic bookshelf insertion and examined how control authority can be divided between geometric structure and learned adaptation. The proposed hybrid controller retains a simple geometric strategy for insertion and post-release seating, while residual PPO provides bounded local corrections and learns when to release the book. This allows learning to focus on the parts of the task that become difficult to specify analytically under contact.

In deployment-matched simulation, the final residual policies achieve \(98.50\%\) mean success compared with \(37.89\%\) for Nominal-only control. On the xArm7, Residual PPO improves end-to-end success from \(26.7\%\) to \(63.3\%\) across 30 matched conditions and reduces the number of failures by half. The paired results show that the improvement persists across different physical constraints and shelf locations rather than arising from a small number of favourable trials.

The remaining failures also clarify the limits of the approach. Performance drops sharply when geometric clearance becomes extremely small, and the Tight physical configurations remain challenging because contact can deform or move neighbouring books and alter the available opening. Residual learning can therefore compensate for moderate modelling and contact uncertainty, but it cannot create geometric freedom that is physically unavailable. Overall, the results support a practical control strategy for contact-rich manipulation: preserve useful geometric task structure and concentrate learning on the local interactions where that structure becomes brittle.
\section*{AI Use Disclosure}
OpenAI ChatGPT and Anthropic Claude assisted with language editing,
manuscript restructuring, and drafting portions of the text. All
technical content and results were verified by the authors.
\FloatBarrier
\bibliographystyle{IEEEtran}
\bibliography{references}
\end{document}